\documentclass[sigconf]{acmart}

\usepackage{booktabs} 

\copyrightyear{2023}
\acmYear{2023}
\setcopyright{rightsretained}
\acmConference[SAC '23]{The 38th ACM/SIGAPP Symposium on Applied Computing}{March 27-March 31, 2023}{Tallinn, Estonia}
\acmBooktitle{The 38th ACM/SIGAPP Symposium on Applied Computing (SAC '23), March 27-March 31, 2023, Tallinn, Estonia}\acmDOI{10.1145/3555776.3577656}
\acmISBN{978-1-4503-9517-5/23/03}

\acmArticle{4}
\acmPrice{15.00}

\begin{document}
\title{Automatic Detection of Signalling Behaviour from Assistance Dogs as they Forecast the Onset of Epileptic Seizures in Humans}

\renewcommand{\shorttitle}{Assistance Dogs for Detecting the Onset of Epileptic Seizures}

\author{Hitesh Raju\textsuperscript{1}, Ankit Sharma\textsuperscript{1}, Aoife Smeaton\textsuperscript{2}, Alan F. Smeaton\textsuperscript{1,3}}

\affiliation{
\textsuperscript{1}School of Computing, Dublin City University, Ireland \\ 
\textsuperscript{2}Dogs for the Disabled, Cork, Ireland\\
\textsuperscript{3}Insight Centre for Data Analytics, Dublin City University, Ireland \\ 
}
\email{alan.smeaton@dcu.ie}

\renewcommand{\shortauthors}{H. Raju et al.}

\begin{abstract}
Epilepsy or the occurrence of epileptic seizures, is one of the world's most well-known neurological disorders affecting millions of people.  Seizures mostly occur due to non-coordinated electrical discharges in the human brain and may cause  damage, including collapse and loss of consciousness. If the onset of a seizure can be forecast then the subject can be placed into a safe environment or position so that self-injury as a result of a collapse can be minimised.
However there are no definitive methods to predict seizures in an everyday, uncontrolled environment. Previous studies have shown that pet dogs have the ability to detect the onset of an epileptic seizure by scenting the characteristic volatile organic compounds exuded through the skin by a subject prior a seizure occurring and there are cases where assistance dogs, trained to scent the onset of a seizure, can signal this to their owner/trainer. 
In this work we  identify how we can automatically detect the signalling behaviours of trained assistance dogs and use this to alert their owner. Using data from an accelerometer worn on the collar of a dog we describe how we gathered movement data from 11 trained dogs for a total of 107 days 
as they exhibited signalling behaviour on command. We present the machine learning techniques used to accurately detect signalling from routine dog behaviour. This work is a step towards automatic alerting of the likely onset of an epileptic seizure from the signalling behaviour of a trained assistance dog. 
\end{abstract}

%
%

\ccsdesc[500]{Applied computing~Life and medical sciences}
\ccsdesc[300]{Hardware~Sensor applications and deployments}
\ccsdesc[300]{Computing methodologies~Supervised learning}

\keywords{Epileptic seizure, seizure-alert dogs, animal signalling behaviour, wearable accelerometer, feature selection, machine learning}

\maketitle

\section{Introduction}

Epilepsy, the occurrence of epileptic seizures, is one of the world's most common neurological disorders affecting millions of people daily.  Seizures  occur due to non-coordinated electrical discharge in the  brain and may cause  damage  including collapse and loss of consciousness  or worse. If a the onset of a seizure can be detected then the subject can be placed into a safe environment or positioned so the risk of self-injury as a result of a collapse can be minimised. However there are no reliable and practical methods to predict the onset of seizures in an everyday, uncontrolled and routine environment.  

Previous studies have shown that  dogs, either household pets or trained assistance dogs, have the ability to forecast the onset of an epileptic seizure by detecting the scent of a particular kind of volatile organic compound associated with seizure onset \cite{MAA2021107690}. Researchers have also demonstrated the presence of seizure-associated odours secreted from the skin and then recording the reactions and signalling behaviour of a pet dog when it senses such odours.  

In the present study we investigate how to automatically detect the signalling behaviour of  trained assistance dogs who  detect a seizure onset and who showcase this through their signalling. 
For this we use a wearable sensor  including an accelerometer on the collar of the dog. We develop a data processing pipeline to pre-process the sensor data and use this as input to a machine learning algorithm that  recognises the assistance dog's signalling behaviour. The accuracy of the detection of the signalling behaviour is such that this one step in the overall goal of using assistance dogs in practice, is demonstrated as being feasible.

The rest of this paper is organised as follows. In the next section we present some background and  literature  including the challenges in this work in Section~\ref{sec:RQs}. We then give an overview of our data processing pipeline followed by a description of the sensor data we  gathered from trained dogs, including the annotation of this as data for training a machine learning algorithm.  In Section~\ref{sec:preproc} we describe our data pre-processing including feature extraction,  data analysis and the algorithms  we used for classification and prediction.  In Section~\ref{sec:results} we present our results including a discussion before our final conclusions.

\section{Background and Literature Review}

\subsection{Epilepsy}
Epilepsy is a neurological brain disorder affecting  millions of people daily. It is characterised by the occurrence of spontaneous and  seizures where the subject loses control of motion and may jerk and thrash around uncontrollably. 
In simple terms a seizure can be defined as an internal electrical storm in the brain. It is the consequence of abnormal, excessive discharges by nerve cells. The seizures in epilepsy are mostly related to a brain injury or are hereditary, but more often the cause is completely unknown \cite{powell2021untrained}. 
Seizures result from a non-coordinated neuronal electrical discharge and recurring seizures occur in the cerebral cortex \cite{9230731} and may cause serious damage, including loss of consciousness and collapses and  self-inflicted injury where  the subject can bang their head on the ground \cite{litt2002prediction}, for example.

Some of the common causes of epilepsy are inborn or developmental including genetic abnormalities or structural problems in the brain like malformed veins or areas which have not developed normally.
%
One in three people will be diagnosed with epilepsy by the age of 60. The unexpected and uncontrolled seizures not only make subjects feel vulnerable but also impact their quality of life.  While there are medical interventions which can reduce the seriousness or the number of seizures, many of these pharmacological interventions have undesired side-effects and getting the dosage as well as the combination of medicines correct, is difficult.

\subsection{Seizure Onset Prediction}

The quality of life of those who experience epileptic seizures would be significantly improved if the onset of seizures could be predicted in advance. In such a case, a  warning mechanism for the onset of a seizure would allow subjects time to remove themselves from a potentially dangerous situation, e.g. from a staircase, to somewhere safer like lying on a bed, a sofa or even lying on the floor. Furthermore, seizure prediction offers the possibility of intervening in the dynamics of the brain before or at the start of  a seizure and ultimately aborting seizures before they occur \cite{7591034}.

Although research has been carried out on predicting epileptic seizures since the 1970s, there are still no  reliable or practical methods available to predict seizure onset in subjects with epilepsy. Therefore there is a pressing need to develop approaches to seizure onset detection. 
The introduction of any technique to detect and accurately predict the onset of a seizure, giving sufficient time for a person to take control of the situation  would also encourage a healthier and more normal lifestyle as the risk of self-industry during a seizure would be reduced. Since seizures are unpredictable, those who are susceptible to seizures must live a lifestyle which avoids activities which contain any serious kind of risk, certainly when they are operating without a carer or assistant \cite{9297423}.

\subsection{Assistance Dogs for Predicting Seizure Onset}


Humans have studied animal behaviour for hundreds of years including during domestication \cite{8998379}.
While most dogs in domestic settings are pets and have evolved to naturally live with human families offering companionship, assistance dogs are trained to undertake a variety of tasks to help individuals. Sometimes these  individuals have  physical or even cognitive disabilities that impair their quality of life.

One form of assistance that assistance dogs can offer is  anticipating the onset of seizures. Early work in this area was reported more than 2 decades ago, work which analyse the reaction of dogs to odours and that work found that the dogs could predict when a seizure was imminent based on smell alone \cite{litt2002prediction}.
This early work was followed  by work which  proved that epileptic seizures are associated with a specific kind of odour which comes from a naturally occurring volatile organic compound that we secrete through our skin, and which is detectable by pet dogs \cite{brown2011can}. 
More recent work by Powell {\em et al.} found that this scent which is an indicator of seizure onset can be detected even by untrained pet dogs \cite{powell2021untrained} while work in  \cite{MAA2021107690}  isolated a unique seizure scent which acts as a biomarker for seizure onset indication.

In addition to the potential for scent-based detection of seizure onset, visual cues from a subject's  behaviour or postural changes  (e.g. changes in motor activity and or mood, or experiencing of auras) prior to an impending seizure can also  be a form of signalling from a subject which could be detected by an assistance dog \cite{catala2018dog}.  Whether scent-based or behaviourally based,  this past work indicates that  assistance  dogs could effectively be trained to recognise a seizure onset and then  give an overt signal by their behaviour, to warn his/her owner or caregiver.

\subsection{Wearable Accelerometers on Animals}

The use of wearable accelerometers by people to measure physical activity, sedentary time,  energy expenditure and sleep-related behaviours, is a well-established market and the billion-dollar pet industry is now getting involved in this wearables market. The global pet wearable market is expected to grow at a compound annual growth rate of 13.5\% before 2025, with enormous growth in demand from Asia after initial popularity in the West. The typical pet wearables on the market are dog activity and fitness trackers, like FitBark, PetPace, and Whistle. These devices employ accelerometers, GPS, and vitality sensors to measure the activity and sleep patterns of pets \cite{8812161}.

Accelerometers are  used in animal science in various contexts such as estimation of energy expenditure and assessment of behaviours in wildlife \cite{kumpulainen2021dog}. Existing activity trackers are useful for evaluating simple canine behaviours according to past research. Depending on the sensitivity of the measurement unit, the spontaneous activities of a dog, such as locomotion, postural change and movement of body in each posture, can be differentiated by analysis of accelerometer data \cite{yam2011validity}. Accelerometer data can be used to extract more detailed behaviours to assess an animal's health and wellbeing. Previously these devices were used to detect musculoskeletal problems, ear infections and mental health conditions such as separation distress \cite{den2017external}. A number of systematic reviews of how to process data from wearables showcases that various attributes must be considered for accurate data collection and processing. These include placement of the sensor, sampling frequency, filters used on the gathered data, epoch length and non-wear-time \cite{migueles2017accelerometer,riaboff2022predicting}.

\subsection{Machine Learning on Accelerometer Data}

The accelerometer data from collar-mounted canine activity monitors can be used to measure a dog's activity levels including step count and distance travelled. Recent advances in machine learning and embedded computing have made it possible to classify animal behaviour in a more nuanced and accurate manner. Through a collar-mounted accelerometer, one study confirmed that activity monitors using validated algorithms can accurately detect important canine behaviours \cite{chambers2021deep}. Validated algorithms have wide practical applications when used in commercially available canine activity monitors. Based on the accelerometer data, algorithms accurately detected eating and drinking behaviour, other behaviours such as licking, petting, rubbing, scratching, and sniffing were also accurately detected \cite{chambers2021deep}. 

In this work we set out to collect information from a number of volunteers  participants and their dogs on  signalling behaviours or ``tricks'' that their dogs can perform while wearing an accelerometer to collect data on  their movement. We also gather information on the various characteristics of the dogs including age, size, breed and the range of signalling behaviours it can perform. Video footage of the dogs performing those behaviours will also be recorded for short instances several times during the data logging period.  When we manually annotate this video footage for the occurrence of such behaviours it gives us a ground truth of positive instances. We can then combine the behaviour detection results across a variety of dog breeds and determine the level of accuracy for automatic detection of seizure onset based on the analysis of accelerometer data from accelerometers worn on the collars of trained assistance dogs.

\subsection{Challenges}
\label{sec:RQs}

In order to investigate how to detect canine signalling behaviour we collect data from  an always-on accelerometer combined with videographic recordings of dogs while performing their signalling behaviour. We use this as training data for a machine learning algorithm. 

Using a triaxial accelerometer comes with hidden challenges. One problem  is that accelerometer data may not be directly comparable across dogs. 
The shape and size of the dog that the device is attached to raises issues such as its relative height with respect to the ground which can alter its relative position and  result in  data  being dissimilar across different  dogs.
Thus the dogs' physical characteristics which include size, age, weight and breed are   important. Dogs which are physically strong and well-built may not be as  agile as lighter breeds of the same size who might perform certain signalling activities more rapidly.

Previous studies have indicated that orientation inconsistency of a wearable sensor is  important  \cite{9141313}. For placement on a dog, the best placement for an activity sensor has been determined to be ventral attachment to the neck collar because this placement also makes it possible to detect behaviours that do not involve movement of the whole body, such as scratching or eating \cite{9774398}.  
However, when worn by participant dogs on their collars, the collar and thus the sensor can rotate around the neck, or be attached initially in  different orientations. 
In some scenarios accelerometers may be mounted on a rigid structure such as a fixed harness worn by the dog in order to maintain a standard sensor orientation. While this addresses the issue of sensor orientation, wearing a harness 24/7 is uncomfortable for an assistance dog and this is not an option for our eventual use case so in this work we  fix the sensor on the dog collar even if the collar may rotate around the dog's neck. 

To isolate any  bias from inconsistency of the sensor  orientation  we  compute the average sum of the three axial values. Thus, for the purpose of our work we  treat rotation of the sensor and the collar as systematic biases that are associated with the devices. A more in-depth and principled analysis of this assumption is beyond the scope of the current study and the reader is referred to \cite{riaboff2022predicting} for further discussion on this topic.

\section{Data Gathering and Annotation}
\label{sec:datagathering}

Data for this work was gathered from a group of 11 volunteer dogs with an age ranging  from 1 to 9 years. All dogs were deemed to be agile and healthy by their owners and this study was approved by the Research Ethics committee of
Dublin City University School of Computing.  All volunteer dog owners read a plain language statement and signed and returned an informed consent form. Volunteers were recruited by information leaflets circulated among  dog training social media groups and almost all volunteer owners were themselves dog trainers.
Information on each volunteer dog was documented including name, the tricks or signalling behaviours that it could perform on command and  we purposely included a range of breeds,  and ages among the volunteer dogs as shown in Table~\ref{tab:dogs}. 
 The unprocessed sensor data is publicly available at~\cite{Smeaton2022}.
In screening volunteer dogs for participation, the ability to perform certain agile  activities such as spins and jumps were  mandatory  and dog owners were informed of this prior to recruitment.

\begin{table*}[ht]
\centering
\caption{Characteristics of Volunteer Dogs}
\label{tab:dogs}
        \begin{tabular}{cllllll}
        \toprule 
       Number & Breed & Size & Age & Data logging & Behaviours & No of videos\\
        \midrule 
       1 & Collie Mix & Medium & 3 Yrs & 9 days & Spins, Jumps and Rollovers & 20\\
       2 & Springer/Terrier MIX & Medium-Small & 8 Yrs & 4 days & Spins, Jumps & 16\\
       3 &  Spaniel  Mix & Medium & 4 Yrs & 9 days & Spins, Jumps & 27\\
       4 &  Poodle & Medium-Small & 2 Yrs & 9 days & Spins, Jumps & 9\\
       5 & Lurcher & Medium & 5 Yrs & 3 days & Spins, Jumps & 14\\
       6 & Lurcher & Medium & 2 Yrs & 9 days & Spins, Jumps & 9\\
       7 & Collie Mix & Medium & 3 Yrs & 8 days & Spins, Jumps & 8\\
       8 & Hungarian Vizla & Medium & 3 Yrs & 12 days & Spins, Jumps & 6\\
       9 & Border Collie & Medium & 2 Yrs 4 Months & 19 days & Spins, Jumps & 11\\
       10 & Australian Retriever & Medium & 1 Yr & 14 days & Spins, Jumps & 9\\
       11 & Golden Retriever & Medium & 7 Months & 11 days & Spins, Jumps & 3\\
       \midrule
       &&&Average age 3.5 Yrs & Total 107 days & & Total 132 \\
        \bottomrule
        \end{tabular}
\end{table*}

\begin{figure}[htb]
\centerline{\includegraphics[scale=.15]{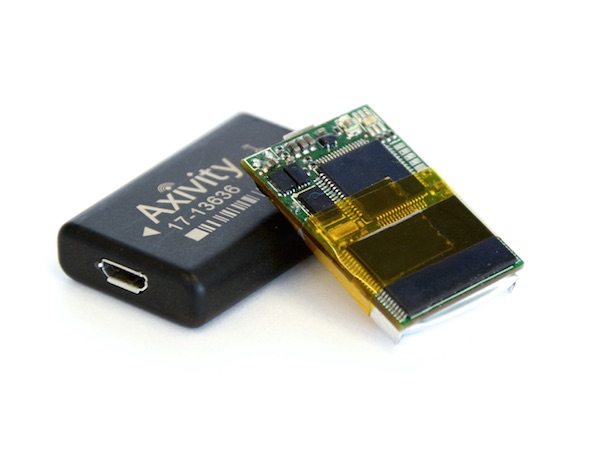}}
\caption{AX3 Accelerometer}
\label{AX3}
\end{figure}

The AX3 accelerometer (Axivity, York, UK \url{https://axivity.com/}) is a popular device for wearable sensing research. A pre-configured and full-charged device with parameters preset for data gathering as described below, was sent to each volunteer and placed securely on the dog's collar and left in place to gather data  in the dog's natural  environment and routine behaviour.  The AX3 data logger, shown in Figure~\ref{AX3}, is a wearable sensor which includes a microelectro-mechanical systems (MEMS) accelerometer and flash memory.  Multidimensional data collection and configurable resolution/frequency are two of the reasons why the AX3 sensor was chosen. This allowed the sensor to be set at  a sampling rate up to 100Hz if needed.

Based on  literature describing the use of the AX3 in other animal studies, a sampling rate of 12.5Hz and 12 bits for data resolution was considered optimal for our experiment \cite{riaboff2022predicting}, trading battery live vs. sensing frequency. It is possible to represent sensor data captured as either single-dimensional data or multi-dimensional data which constitutes the x, y, and z directional dimensions. The energy consumption for the wearable device is an important parameter due to the fact that dog activity should be tracked over a prolonged period of time. Reducing sample rates, logging fewer accelerometer axes, or reducing features to record can increase battery life and for our configuration, battery life for full charges is between 6 and 8 weeks. A recent study examined dairy cows' laying, standing, and feeding behaviour and found that fewer logging axes did not affect the accuracy of the categorisation \cite{vazquez2015classification}.
By choosing  optimal sensor measurement settings such as sampling rate, logging axis, and feature selection, the energy consumption in a wearable device  can be reduced significantly. A number of studies have examined these features of accelerometer data for human activity classification, but none have examined how they affect behaviour recognition accuracy in dog tracking.

\begin{figure}[htb]
\centering
\includegraphics[width=\columnwidth]{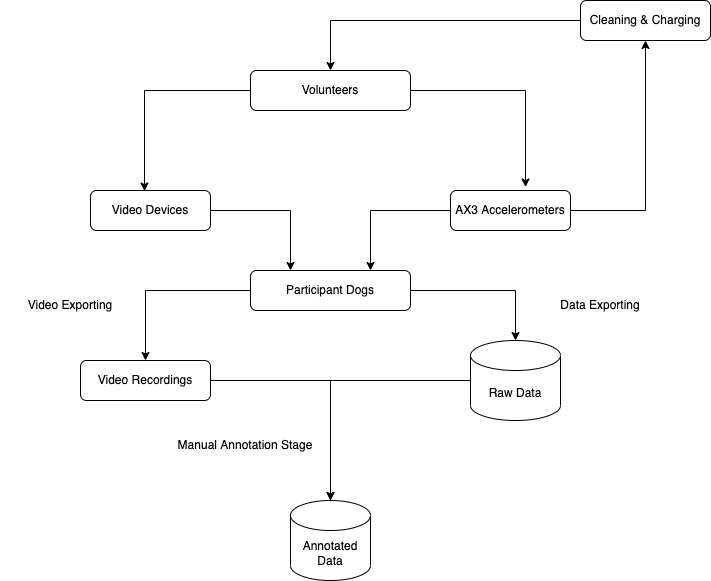}
\caption{Data flow during data gathering}
\label{Architecture}
\end{figure}

Each dog performs a range of different  activities as part of their natural behaviour during the data logging period including standing, sitting, sleeping, lying, jumping and running, and all the this while wearing an accelerometer on their collar. Occasionally during the data logging period and at a time of her/his choice, the owner would record a video on a smartphone of their dog performing a range of signalling behaviours on command. A total of 132 such videos were  shared with us  through an online platform as soon as the video recordings were made. This allowed us to retrospectively identify and synchronise the video recordings with the accelerometer data. When the logging period was completed and the sensor returned to us, data was downloaded and the device was cleaned and shared with the next participant to repeat the  process.  Figure~\ref{Architecture}  outlines the  data flow of the data gathering process. 

Table~\ref{tab:activity}  shows the  most common of the video-recorded dog behaviours in our investigation across all dogs indicating the activity performed and the  time to complete it as averaged across all  participant dogs.  Not all dogs were able to complete all behaviours which meant we had to decide on a single signalling behaviour which was 
\begin{enumerate}
    \item common across many/most/all assistance dogs or that an assistance dog could be trained to do;
    \item rarely occurring in a dog's natural behaviour, so not sitting for example as that is a fairly commonplace activity;
    \item distinct enough in terms of motion characteristics to make it more easily distinguishable from other dog movements.
\end{enumerate}

\begin{table}[ht]
\centering
\caption{Activity Table}
\label{tab:activity}
        \begin{tabular}{clc}
        \toprule 
        Number & Activity type & Average completion time (all dogs)\\
        \midrule 
        1 & Spin clockwise & 1.5 sec \\
        2 & Spin anti-clockwise & 1.2 sec \\
        3 & Stand & 2.2 sec \\
        4 & Jump & 1 sec \\
        5 & Sit & 1 sec  \\
        6 & Rollover & 2 sec \\
        7 & Idle & N/A \\
        8 & Other Movements & N/A \\
        \bottomrule
        \end{tabular}
\end{table}

\begin{figure}[htb]
\centerline{\includegraphics[width=0.8\linewidth]{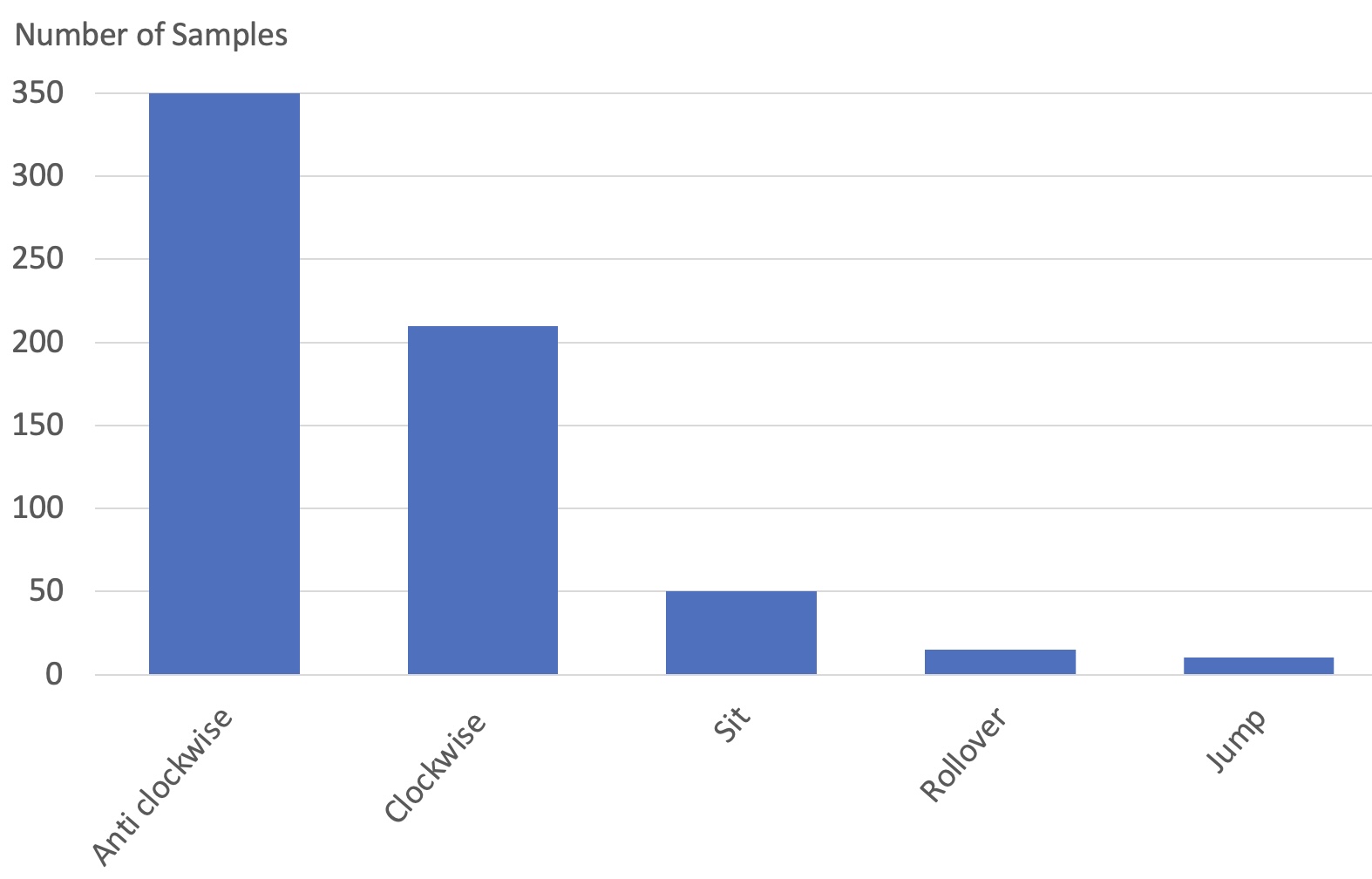}}
\caption{Numbers for each signalling behaviour manually annotated as training data in volunteer's videos}
\label{Samples}
\end{figure}

Behaviours observed during the recorded sessions across all dogs included clockwise spins, anti-clockwise spins,  sitting, standing, jumping and rollovers as indicated in Figure~\ref{Samples}. From this we can see that spins, where a dog spins in a circle with all four legs on the ground, is a signalling behaviour that meets our requirements of being easy to train, occurs rarely in day-to-day activities and is distinct enough from other dog movements to be automatically detected. There is also a sufficient number of example recordings in the videos to provide enough data on which to train a machine learning algorithm to recognise this behaviour automatically.  

The  data recorded on the AX3 accelerometers was downloaded at the end of each particiant's logging period when the sensor was returned to us and the raw CWA-file was exported  for further processing. Data was validated to confirm that the actual sample rate of the accelerometers matched the sample rate predefined in the OMGUI software (\url{https://github.com/digitalinteraction/openmovement/wiki/AX3-GUI}) used to configure and manage the AX3. The recorded data was then trimmed to remove data logging while the AX3 was in transit from us to the dog (by post) and from the participant back to us (by post)  and cases where it was idle for prolonged periods. The volunteers had been briefed prior to sensor placement about the range of behaviours that should be performed and recorded, and the raw sensor data was manually inspected in order to annotate it with a corresponding activity label by viewing the synchronously  recorded videos.
This was also the method used to annotate sensor data. 

As part of the process of annotating video data, activity definitions were set before analysing the videos.  The accelerometer signal was not visible to the annotators while annotating. Video recordings were viewed locally and manually annotated frame-by-frame by two observers blinded to the sensor data and using a controlled vocabulary. To calculate the annotation format, the initial and final times of each activity were recorded. All the annotated information was saved  which included the number and labels of movements and their time intervals, each label corresponding to a single dog behaviour. Sensor data was  imported into CSV format and synchronised with the videos for each dog. Once annotated, data from each dog’s recording session was exported into a separate data file containing features for time, sensor data, and the annotated signalling behaviour.

\begin{figure}[htb]
\centerline{\includegraphics[scale=0.3]{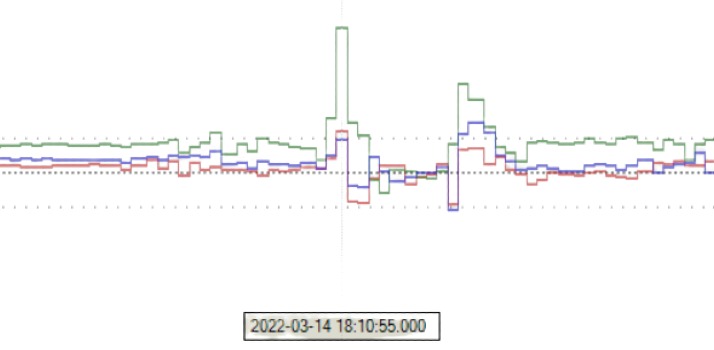}}
\caption{Jump indication on the y-Axis}
\label{JumpY}
\end{figure}

\begin{figure}[htb]
\centerline{\includegraphics[scale=0.3]{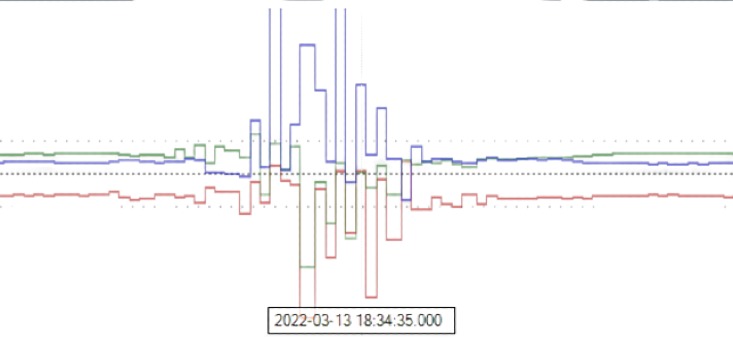}}
\caption{Jump indication on the x-axis and z-axis}
\label{JumpXZ}
\end{figure}

Several important points were identified during the annotation stage where sensor orientation inconsistency was observed and noted. Several examples of orientation inconsistencies were identified. In  Figure~\ref{JumpY} for example we see spikes in the intensity of motion in the y-axis and  to a much lesser extent in the  x- and z-axes of the accelerometer (blue and red lines respectively in Figure~\ref{JumpY}) while the  jump indications were observed to a much higher extent  in the accelerometer's y-axis (green line in Figure~\ref{JumpY}. Yet for the same dog but recorded a day earlier, the recordings for the same jumping behaviour is shown in Figure~\ref{JumpXZ} indicating the detection of the jumping motion by the x-axis (blue line in Figure~\ref{JumpXZ}) and inversely by the z-axis (red line in Figure~\ref{JumpXZ}) as well as by the y-axis (green line in Figure~\ref{JumpXZ}). This indicates that for the same jumping behaviour by the same dog and of approximately the same intensity (through the graphs in Figures~\ref{JumpY} and \ref{JumpXZ} are not to the same scale), the dog's collar must have rotated around her neck and hence the sensor's orientation had changed. This demonstrates the challenge of managing the inconsistency of sensor orientation. 

Finally the common annotations from the two observers were merged and exported to a single file. This method was followed by annotating for behaviour for each dog individually, by each annotator independently.  Where annotators disagreed, i.e. where one annotator indicated a signalling behaviour and the other did not, then that annotation was discarded and it was only the commonly-agreed annotations that were retained. Once the annotation process was completed, the frame-by-frame output and time labels were exported to another CSV file. Finally this data could be used in various combinations to carry out the investigations.  The number of annotations for each of the major dog activities is shown  in Figure~\ref{Samples}.

\section{Data Preprocessing}
\label{sec:preproc}

Before developing algorithms for the classification of dog signalling behaviour, it is important to understand the nature of the signals produced by the triaxial accelerometer. These consist of three separate data streams that represent time series for acceleration on each axis  x, y and z. Complementary to the three axes, an additional time series, the named sum can been obtained by computing the magnitude of the acceleration described   into signal vector magnitude (SVM), a time-series independent of  sensor orientation and  invariant to any movement of the collar around the dog's neck.  SVM is also known as {\it Amag} and shown below, is a useful metric for calculating movements which are not axis-specific and is used extensively in pre-processing raw accelerometer data \cite{riaboff2022predicting}.  

\[ Amag = \sqrt{x^2 + y^2 + z^2}  \]

\begin{figure}[htb]
\centerline{\includegraphics[width=\linewidth]{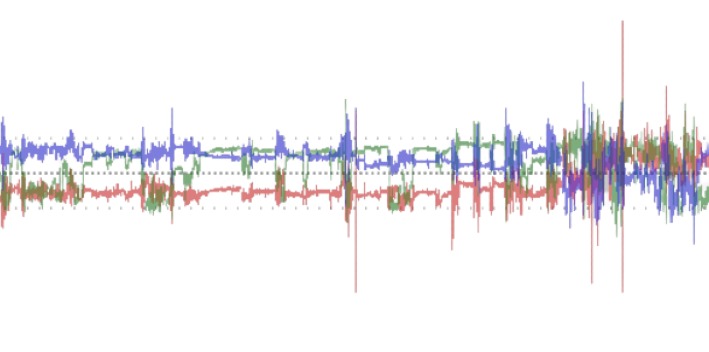}}
\centerline{\includegraphics[width=\linewidth]{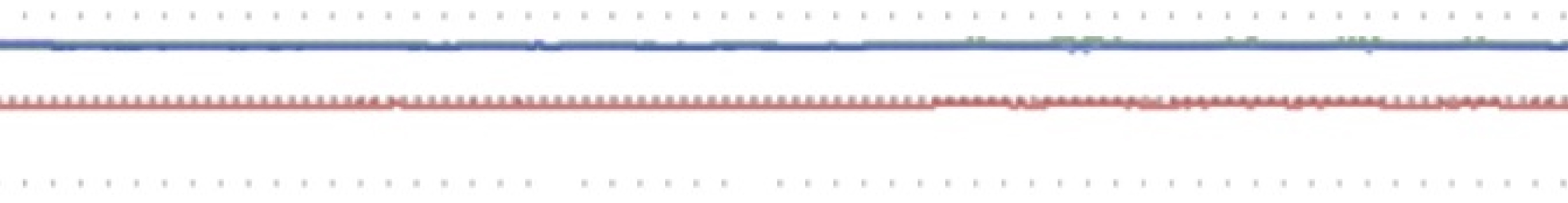}}
\caption{Sample recordings of dog activity. The top graph shows movement in all 3 directions while the bottom graph shows a dog at rest with no movement.}
\label{activity}
\end{figure}

Some examples of raw accelerometer recordings are in Figure~\ref{activity} indicating indicates a variety of movements in all directions being captured on the sensor in the top graph and  an idle stage without any movements on the sensor indicating the dog is still in the bottom graph.   Raw accelerometer data was recorded for all the participant dogs at 12Hz, giving 12 values of raw accelerometer data for each second. In order to obtain the most important and discriminative statistical features of the data we applied feature engineering where the sensor data was used to derive several features. An important advantage of feature engineering is that it allows the most relevant and important features to be extracted from a collection of data. Our feature extraction process is outlined  below.

\subsection{Feature Extraction}
Feature sets are  reduced sets of original data that represent main characteristics and behaviours and are thus abstractions of raw data. A feature vector is a subset of larger input data that contains important clues for activity recognition and is used as an input into clustering or classification algorithms. In our work we used the Time Series Feature Extraction Library (TSFEL) written in Python which automatically extracts  over 500 features from a time series \cite{barandas2020tsfel}. 
These include Fast Fourier Transform (FFT), Wavelet Transform (WT), Autocorrelation, Mean, Median, Sum of Absolute Differences, Entropy, Peak to Peak Distance, Slope, Zero crossing rate, Interquartile range, ECDF percentile count and ECDF slope, Spectral maximum peaks, Maximum power spectrum, etc. 
TSFEL can handle multidimensional time series, as is the case here.

In feature selection it is advantageous to use smaller feature sets that are fast to compute when there is a shortage of computational resources in the applications. While working with larger sets of complex time-series features can be slower to compute, it can result in higher performance in downstream applications where accuracy is more important. Feature extraction is usually applied to windows taken from a time series and testing is required to determine how the signal window size affects subsequent classification performance.

\subsection{Relevant Features}
The characteristics of a given classification task may require different features to be extracted from data, and some variables in the original data may be irrelevant or redundant. A feature selection technique is used to identify the most informative features and to limit the computational demands when applying the recognition system to new observations. The feature selection stage selects a smaller subset of the original features to identify those most useful features. The choice of features acquired from a data set and the window length over which they are computed are key factors. 

In the work addressed here, recognising dog signalling  behaviour depends directly on features extracted for motion analysis. The same signalling behaviour movement can be performed by different dogs in  different ways depending on their size, speed, and age, resulting in considerable variability in the features derived from the data collected by collar-worn sensors. It is therefore important  to identify features with high discriminative ability in order to achieve effective classification. An effective feature set should show little variation between repetitions of the same movements performed by the same individual dog, but should differ between different dogs performing the same task. To achieve this we   extract only features  related to motion and energy. This will enhance the accuracy of predicting a movement across a variety of sensor readings. 

A dictionary of features is developed which categorises them into three types: statistical, temporal, and spectral as indicated in Table~\ref{tab:Features}. Some of these have been extensively investigated in previous studies and have proven useful in recognising activity and motion \cite{vazquez2015classification}. A window size of 12 data values was  pre-selected to correspond to approximately 1 second duration since  the frequency in the AX3 is 12.5Hz, and a window overlap of 50\% was set for our feature extraction. 

\begin{table}[ht]
\centering
\caption{Motion features for classification from time series}
\label{tab:Features}
        \begin{tabular}{p{2.5cm}lp{2.5cm}}
        \toprule 
        Statistical features & Temporal features & Spectral features\\
        \midrule 
        Kurtosis; & Zero crossing; & Maximun frequency; \\
        Skewness; & Pairwise correlation; & Median frequency; \\
        Mean; & Auto Correlation; & Cepstral Coefficients; \\
        Standard Deviation; &  & Power Spectrum; \\
        Interquartile Range; &  & Power Bandwidth;\\
        Root Mean Square; &  & Fundamental \\
        Median Absolute Deviation; &  &Frequency Spectrum; \\
        \bottomrule
        \end{tabular}
\end{table}

The final step in data preprocessing and feature extraction is to address the  difference in magnitude among the features obtained from TSFEL as several machine learning algorithms emphasise more heavily those features with a higher magnitude \cite{barandas2020tsfel}. A normalising step is thus applied to features before classification to prevent this unwanted effect. To account for different scale factors and units, all described features are normalised to zero mean and unit variance before proceeding to algorithm development and analysis.

\subsection{Algorithm Development}
\label{sec:dataanalysis}

In order to verify if the selected features selected are  informative, we use different classification methods for  various dog signalling behaviours. Because the optimisation criteria are convex, Support Vector Machine classifiers are desirable in this scenario. The support vector machine is a linear classifier based on supervised learning, but it can also be used as a nonlinear classifier by using different kernel functions \cite{cervantes2020comprehensive}. In Support Vector Machine learning, the basic idea is to find the separation hyperplane that can divide the training data set correctly and have the largest set intervals. We also trained and tested the data set on other classification algorithms including logistic regression, KNN, Random forest and Naive Bayes. All our results were validated by 10-fold cross validation.

Our experiments were conducted using the Python programming language and the interface was Jupyter Labs. A standard M1 processor Macintosh with 16GB of CPU memory was used and  processing time varied across different classification models.

\section{Experimental Results}
\label{sec:results}

In prediction and classification of assistance dog signalling behaviour, conventional machine learning algorithms based on accelerometer data can be used. To distinguish between different dog behaviours and features extracted, supervised learning is used in this work. Due to the availability of annotated training data, supervised approaches are thus straightforward to implement. 

In our experiments we ran different machine learning classification algorithms  on the features extracted from  accelerometer data from the 11 assistance dogs. The algorithms we used included  Support Vector Machines, Logistic Regression, kNN, Random Forest and Naive Bayes. A classifier for each of  the 11 participant dogs was trained and tested on data from that dog only, using a support vector machine classification model and the results for the detection of dog spins (in either direction) are indicated in Table~\ref{tab:classification-perdog}.

\begin{table*}[ht]
\centering 
\caption{Support vector machine classification results for individual classifiers for 11 participant dogs
\label{tab:classification-perdog}}
        \begin{tabular}{clccccc}
        \toprule 
        Number & Dog Breed  & Precision & Recall & F1 Score & Support & Accuracy\\
        \midrule 
     1 & Collie Mix  & 0.67 & 0.71 & 0.69 & 17 & 0.78\\
      2 & Springer/Terrier Mix  & 0.67 & 1.00 & 0.8 & 4 & 0.98\\
      3 & Spaniel  & 0.46 & 0.5 & 0.55 & 11 & 0.52\\
      4 & Poodle  & 0.38 & 0.56 & 0.45 & 9 & 0.65\\
     5 & Lurcher  & 0.74 & 0.8 & 0.77 & 14 & 0.79\\
      6 & Lurcher  & 0.17 & 0.5 & 0.25 & 2 & 0.7\\
      7 & Collie Mix  & 0.64 & 0.5 & 0.56 & 14 & 0.65\\
    8 & Hungarian Vizla  & 1.00 & 0.77 & 0.87 & 13 & 0.77\\
    9 & Border Collie  & 0.67 & 1.00 & 0.8 & 2 & 0.94\\
      10 & Australian Retriever  & 1 & 0.93 & 0.97 & 15 & 0.99\\ 
      11 & Golden Retriever  & 0.75 & 0.43 & 0.55 & 7 & 0.76\\
      \midrule 
      &Average & 0.64 & 0.73 & 0.67 && 0.78\\
        \bottomrule
        \end{tabular}
\end{table*}

The results in Table~\ref{tab:classification-perdog} show a range of performances depending on the dog.  Of the 4 evaluation measures included, recall is  the most important for our use case.  When an assistance dog detects a seizure onset from its owner and exhibits signalling behaviour, we always want that behaviour to be detected. Our use case  can tolerate false positives where the algorithm incorrectly interprets a behaviour from the sensor data as being a signal because we can discard a false alert, but we do not want to miss a signalling behaviour, either true or false, in case it is true.  Recall scores range from 0.5 to 1.0 across different dogs with an average of0.73 and while some of the per dog  performance figures may be disappointing we are not interested in training a machine learning classifier to recognise signalling behaviour from each dog independently, we want to pool the annotations to train a single classifier that operates across all dogs. Nevertheless, the performance of the classifier for some dogs, notably dogs numbered 2 (Springer/Terrier), 9 (Border Collie) and 10 (Australian Retriever) was good. All of these are at the larger end of the medium sized breeds with faster movements.

The combined training data of all 11 participant dogs was used to train and test a classifier for dog spinning behaviour in either direction using the set of 5   classification model techniques mentioned earlier. Validation tests were performed using 10-fold cross validation algorithms individually for each participant and with the combined training data. 
%
The results based on the combined training data from all 11 participants  are indicated in Table-\ref{tab:classification} using 5 different machine learning algorithms. Our best overall results were achieved using the random forest  and the naive Bayes algorithms with naive Bayes having the best recall score at 0.98. Performance results for the support vector machine, logistic regression and kNN were not as good.  It should be noted that these results were also calculated using an 10-fold cross validation to further strengthen their validity and demonstrate an acceptable level of performance for this task when used in real world scenarios.


\begin{table*}[!ht]
\centering
\caption{Classification Results for classifiers using training data taken from all dogs
\label{tab:classification}}
        \begin{tabular}{clccccc}
        \toprule 
        Number & Classifier & Precision & Recall & F1 Score & Support & Accuracy \\
        \midrule 
        1 & Support vector machine & 0.84 & 0.62 & 0.71 & 52 & 0.71 \\
        2 & Logistic Regression & 0.74 & 0.67 & 0.70 & 50 & 0.70 \\
        3 & K-NN & 0.33 & 0.08 & 0.12 & 52 & 0.84  \\
        4 & Random Forest & 0.95 & 0.77 & 0.85 & 52 & 0.96  \\
        5 & Naive Bayes & 0.95 & 0.98 & 0.97 & 52 & 0.98  \\
        \bottomrule
        \end{tabular}
\end{table*}

\section{Conclusions}
The combination of data from wearable sensors and  machine learning is an exciting new frontier and opportunity.  The contribution of this work is the design and evaluation of a behaviour recognition system based  on data from a wearable accelerometer that identifies the presence of a pre-trained signalling  behaviour. The set of  features automatically extracted from the sensor data  ensures acceptable recall in terms of classification results. The  results presented here suggest that using a realtime equivalent of the AX3 accelerometer can  identify dogs' signalling behaviour and for an assistance dog trained to alert when sensing the onset of an epileptic seizure, this can then trigger an alert to the owner or carer.

One  possibility suggested to further improve the quality of the classification is to use additional wearable sensors besides accelerometry. \cite{arablouei2022multi} has worked with accelerometry fused with GNSS or GPS data but this or any other form of location information would not be useful in our use case since an assistance dog's signalling behaviour could occur anywhere, including indoors. 

Future research is needed to continue to develop and refine the model to identify additional signalling behaviours that could be performed easily by dogs and for which they could easily be trained. These could include rollovers, beg-ups, and other distinct behaviours. 


\bibliographystyle{ACM-Reference-Format}
\bibliography{ref} 

\end{document}